\documentclass[11pt, oneside, a4paper]{article}
\usepackage[utf8]{inputenc} 
\usepackage[english]{babel} 
\usepackage{graphicx}          
\usepackage{cite}              
\usepackage{enumitem}
\usepackage{pavt-en} 
\usepackage{tabularx} 
\usepackage{ragged2e} 

\begin{document}


\title{ Enhancing Admission Inquiry Responses with Fine-Tuned Models and Retrieval-Augmented Generation }

\authors{Virabyan Aram}

\begin{abstract}
University admissions offices face the significant challenge of managing high volumes of inquiries efficiently while maintaining response quality, which critically impacts prospective students' perceptions. This paper addresses the issues of response time and information accuracy by proposing an AI system integrating a fine-tuned language model with Retrieval-Augmented Generation (RAG). While RAG effectively retrieves relevant information from large datasets, its performance in narrow, complex domains like university admissions can be limited without adaptation, potentially leading to contextually inadequate responses due to the intricate rules and specific details involved. To overcome this, we fine-tuned the model on a curated dataset specific to admissions processes, enhancing its ability to interpret RAG-provided data accurately and generate domain-relevant outputs. This hybrid approach leverages RAG's ability to access up-to-date information and fine-tuning's capacity to embed nuanced domain understanding. We further explored optimization strategies for the response generation logic, experimenting with settings to balance response quality and speed, aiming for consistently high-quality outputs that meet the specific requirements of admissions communications.
\end{abstract}



\section{Introduction}
University admissions offices receive a vast number of inquiries from applicants daily. The quality of responses to these inquiries plays a key role in forming the first impression of the university. Responses from the admissions office are often the first interaction of a prospective student with the university. According to research \cite{cite1}, customer satisfaction directly affects retention, word-of-mouth, and switching intentions. Therefore, it is essential that each response be of the highest quality and professionalism. This task is challenging initially and becomes increasingly difficult as inquiry volume grows, often forcing staff to rush and dedicate less time per response.

\par 
Over the past couple of years, due to the explosion of generative artificial intelligence, more and more use cases are leaning towards large language models (LLMs). Large language models can potentially generate draft responses that admissions staff can review, edit if necessary, and send the response. While simpler rule-based chat bots and similar systems have seen application in educational contexts, they typically lack the flexibility and deep natural language understanding necessary to handle the diverse and complex inquiries common to university admissions offices. This challenge is particularly pronounced in environments like Russia, where admission rules are intricate, heavily regulated by the government, and filled with nuances that often overwhelm basic systems. Our approach, leveraging recent advancements in LLMs through the combination of retrieval-augmented generation and fine-tuning, offers a substantially more robust and scalable solution compared to these earlier methods or reliance on basic prompt engineering alone.

\par 
During the admission campaign, the admissions office of the Higher School of Economics in Russia receives a large volume of inquiries - more than 2,000 per day. This volume often exceeds the capacity of the staff, resulting in a significant increase in response time. Therefore, the time available for each response is shortened, which can prevent staff from providing comprehensive and detailed answers, potentially hindering their ability to fully demonstrate institutional interest in the applicant.

\par 
Large language models can help admissions offices improve the quality of responses to applicants through automation. However, a limitation of mass language models from providers such as ChatGPT, Yandex GPT, GigaChat, and others is that they are not specifically trained on university admissions policies. Consequently, these models lack a comprehensive understanding of university admissions procedures and policies characteristic of experienced admissions staff. For the model to generate high-quality responses, it must develop knowledge and analytical capabilities related to university admissions campaigns. To embed domain-specific knowledge and develop analytical skills in the context of university admissions, we propose a hybrid approach combining Retrieval-Augmented Generation (RAG) with Large Language Model (LLM) fine-tuning. While many studies show that fine-tuning alone is not sufficient for effective knowledge adoption - for example, a \cite{cite2} study concludes that RAG outperforms unsupervised fine-tuning in knowledge-related tasks - RAG has a limitation. In general, RAG functions as a search engine, and its performance can be highly dependent on the specific wording of user queries. This dependency can sometimes lead to suboptimal information retrieval.

\section{Related Work And Theoretical Background}

\subsection{Retrieval-Augmented Generation (RAG)}
In the context of university admissions, RAG has the potential to significantly improve response accuracy. As admissions-related inquiries are highly specific and varied, relying solely on pre-trained models can lead to vague, generalized, or even misleading responses (e.g., confusing information between universities). RAG’s ability to fetch domain-specific documents, such as admissions rules, educational program catalogs, and other relevant documents and data such as specific deadlines, enables the system to generate highly tailored responses. Given the complexity and variability of the admissions process, where details such as application deadlines, program prerequisites, and eligibility criteria are key, RAG becomes particularly effective. By drawing from these specialized documents, the model can generate answers that are not generic, but instead directly relevant to the specific needs of prospective students or applicants. This is a significant advantage in environments where precision and correctness are paramount.

\par 
A recent study \cite{cite3} has indicated that Retrieval-Augmented Generation (RAG) has emerged as a promising paradigm to address the aforementioned challenges. Specifically, RAG introduces an information retrieval process, which has been found to enhance the generation process by retrieving relevant information/documents from available data stores. This, in turn, has led to higher accuracy and greater robustness.

\par 
While various RAG implementations exist, in general, RAG functions by retrieving relevant documents from external knowledge bases, providing accurate, real-time, domain-specific context to LLMs \cite{cite4}.

\subsection{Fine-Tuning in AI Models}
Fine-tuning has become a fundamental technique in the optimization of AI models, particularly when applying general-purpose models to domain-specific tasks. At its core, fine-tuning involves taking a pre-trained model, which has learned general patterns from large datasets, and further training it on a smaller, domain-specific dataset. Recent research \cite{cite6} implies that additional training allows the model to adapt its knowledge to better handle specialized terminology, nuances, and context relevant to the specific domain, improving its performance in tasks such as generating responses to admissions inquiries.

\par 
Fine-tuning is particularly beneficial in situations where a model’s pre-existing knowledge, obtained from a broad and general dataset, may not be sufficient to understand the intricacies of specialized areas. For example, in the context of university admissions, the model must understand complex terminology and procedures. These terms may not be sufficiently represented in general corpora, but they are essential for accurate and relevant responses in an admissions context. Through fine-tuning, the model can be retrained to focus on this specialized vocabulary, improving its ability to generate precise and contextually appropriate answers.

\par 
Research has shown that fine-tuning can lead to substantial improvements in both accuracy and relevance. For instance, Radford et al. \cite{cite7} demonstrated that fine-tuning GPT models on specialized data significantly enhanced their ability to generate relevant, domain-specific text. Similarly, studies in the medical and legal fields have shown that fine-tuned models outperform generic models in understanding and responding to specialized queries \cite{cite2}. The improvements observed in these domains are directly applicable to the university admissions process, where accurate, timely, and contextually aware responses are essential.

\par 
Moreover, fine-tuning is often an iterative process. It requires careful selection of a representative dataset and continuous evaluation to ensure that the model's performance improves over time. In our case, the dataset would consist of a comprehensive collection of admissions-related documents, including policies, program descriptions, application guidelines, and anonymized previous student inquiries. By fine-tuning the model on this dataset, it becomes more adept at generating responses that align with the institution's specific requirements and regulations.

\subsection{Combining RAG with Fine-Tuning}
The combination of RAG with fine-tuning offers the potential to significantly improve response speed and accuracy. By retrieving relevant documents in real-time, the model ensures that the information provided is up-to-date and tailored to the query. Fine-tuning further enhances this process by allowing the model to generate responses that are not only based on the retrieved data but also reflect the specific language and protocols of the admissions process. This results in responses that are not only fast but also precise, reducing the need for manual intervention and follow-up inquiries.

\par 
Furthermore, fine-tuning can help address one of the key limitations of RAG: the need to interpret and synthesize large volumes of information. In a typical RAG implementation, the retrieved documents can be lengthy or contain multiple pieces of information that need to be distilled into a concise and coherent response. Fine-tuning enables the model to better understand how to prioritize and structure this information, ensuring that the response is clear and relevant to the specific needs of the query.

\par 
In practice, the integration of RAG and fine-tuning requires careful calibration. The model must be fine-tuned on a diverse dataset that reflects the variety of inquiries encountered in the admissions process. This may include questions related to specific programs, application procedures, deadlines, and eligibility criteria. The fine-tuning process must also be iterative, with continuous evaluation to ensure that the model is learning effectively and that the quality of its responses improves over time. Additionally, regular updates to the knowledge base may be necessary to ensure that the system remains current with changes in admissions policies and procedures.

\par 
The combination of RAG with fine-tuning also enables the model to handle a broader range of inquiries. While RAG helps the system retrieve relevant data, fine-tuning ensures that the system is equipped to handle complex, domain-specific queries that may not be fully addressed by a general-purpose language model. For example, a prospective student may inquire about specific requirements for an international transfer student, which requires knowledge of both program prerequisites and visa policies. Fine-tuning on a dataset that includes these specifics allows the system to generate a response that is both accurate and comprehensive.

\section{Methodology}

\subsection{Data Collection \& Preprocessing}
The data collection process involves the collection of data related to university admissions, including official documents, data from the admissions website, and previous staff responses (often in a question-and-answer format) extracted from ticketing systems. The collected data is then verified for quality and processed. This includes cleaning the data, removing irrelevant and sensitive information, and normalizing text for consistency. The processed data is stored in a structured format for efficient retrieval and is used to provide a basis for model training.

\par 
The data collection process involves the collection of data related to university admissions, including official documents such as admissions regulatory documents and data from official Q/A pages on the website. The collected data is then verified for quality and processed, which is achieved by fact-checking done by admissions staff. This includes cleaning the data, removing irrelevant and sensitive information, and normalizing text for consistency. The knowledge base is formed with the collected data described earlier.

\par 
At the same time, directly using these raw source materials, especially for fine-tuning, presents significant challenges. For example, official regulatory documents often have complicated structures; maybe section 4.5 refers back to a definition in section 2.9, which is hard for a model to follow directly. Web pages can also be messy, with inconsistent formatting or navigation elements mixed in. Trying to teach the model directly from this kind of unstructured, complex data isn't very efficient and might lead to weird hallucinations and mistakes in understanding; it can struggle to extract the core information or understand the relationships between different pieces of text.

\par 
To address this problem and create a cleaner and more efficient dataset for fine-tuning, we used a data distillation process. We used DeepSeek R1 LLM model \cite{citepre8}, a large language model characterized by its ability to handle extensive context. We provided DeepSeek R1 with collected source documents - regulations, web content, FAQs - as context. We then asked the model to generate relevant question-answer pairs based only on the information contained in these particular documents.

\par 
This distillation procedure resulted in a structured Alpaca-formatted dataset containing about [~8,000] high quality pairs. This effectively transformed the complex, unstructured raw information into clear question and answer pairs that are much more suitable for controlled fine-tuning than the original raw documents. The raw documents themselves were indexed and used primarily for the searchable database of the RAG component.

\subsection{Model Architecture}
For this project, we leveraged a RAG architecture, which combines the power of a large pre-trained language model with an information retrieval system. The architecture consists of two primary components: the retriever and the generator.

\par 
A) Retriever: The retriever is responsible for fetching relevant documents or passages from an external knowledge base. In our case, the knowledge base consists of the domain-specific admissions documents collected during the data collection phase. The retriever uses an embedding-based similarity search approach (specifically, BERT embeddings) to find the most relevant documents that may contain the answers to a given query.

\par 
Documents in the knowledge base were segmented into chunks of [512] tokens with an overlap of [64] tokens to ensure semantic continuity. During retrieval, the [top 3] most relevant document chunks based on cosine similarity with the query embedding were selected.

\par 
B) Generator: The generator is a fine-tuned version of a selected pre-trained language model (Gemma 9B model \cite{cite8} as the base one), adapted using [Low-Rank Adaptation (LoRA)] for [3] epochs on our distilled dataset, responsible for generating natural language responses based on the information retrieved by the retriever. The generator's job is to process the input request and relevant documents provided by the retriever, synthesizing this information into a consistent, accurate, and context-aware response.

\section{Experimental Setup and Evaluation}

\subsection{Experiment Design}

The experiment is structured to compare four different response generation approaches:

\begin{enumerate}
    \item \textbf{Baseline GPT Model}: A standard pre-trained language model without fine-tuning or RAG.
    \item \textbf{RAG Model}: A model incorporating retrieval-augmented generation but without fine-tuning.
    \item \textbf{Fine-tuned model}: A model fine-tuned on domain-specific data but *without* RAG. 
    \item \textbf{Fine-tuned Model with RAG}: Our proposed approach, combining RAG with domain-specific fine-tuning.
\end{enumerate}

The dataset used for evaluation consists of 210 university admissions inquiries collected from historical records. Each model generated responses, which were then manually assessed for accuracy and completeness. Selected inquiries were sampled from historical records to represent a diverse range of common topics, including application procedures, deadlines, document requirements, international student queries, and program-specific questions. Manual assessment was conducted by two team members following predefined guidelines focusing on factual correctness, completeness, relevance, and adherence to university tone. The main factor was whether or not the reviewing admissions officer would have sent the generated response to the author of the appeal (where 0 - would not have sent in any form, 1 - would have sent with some edits, 2 - would have sent without edits). To determine the quality of the completed assessment, inter-rater reliability score is calculated, resulting in: Cohen's Kappa > 0.8.

\begin{figure}[h!] 
    \centering
    \includegraphics[width=0.7\linewidth]{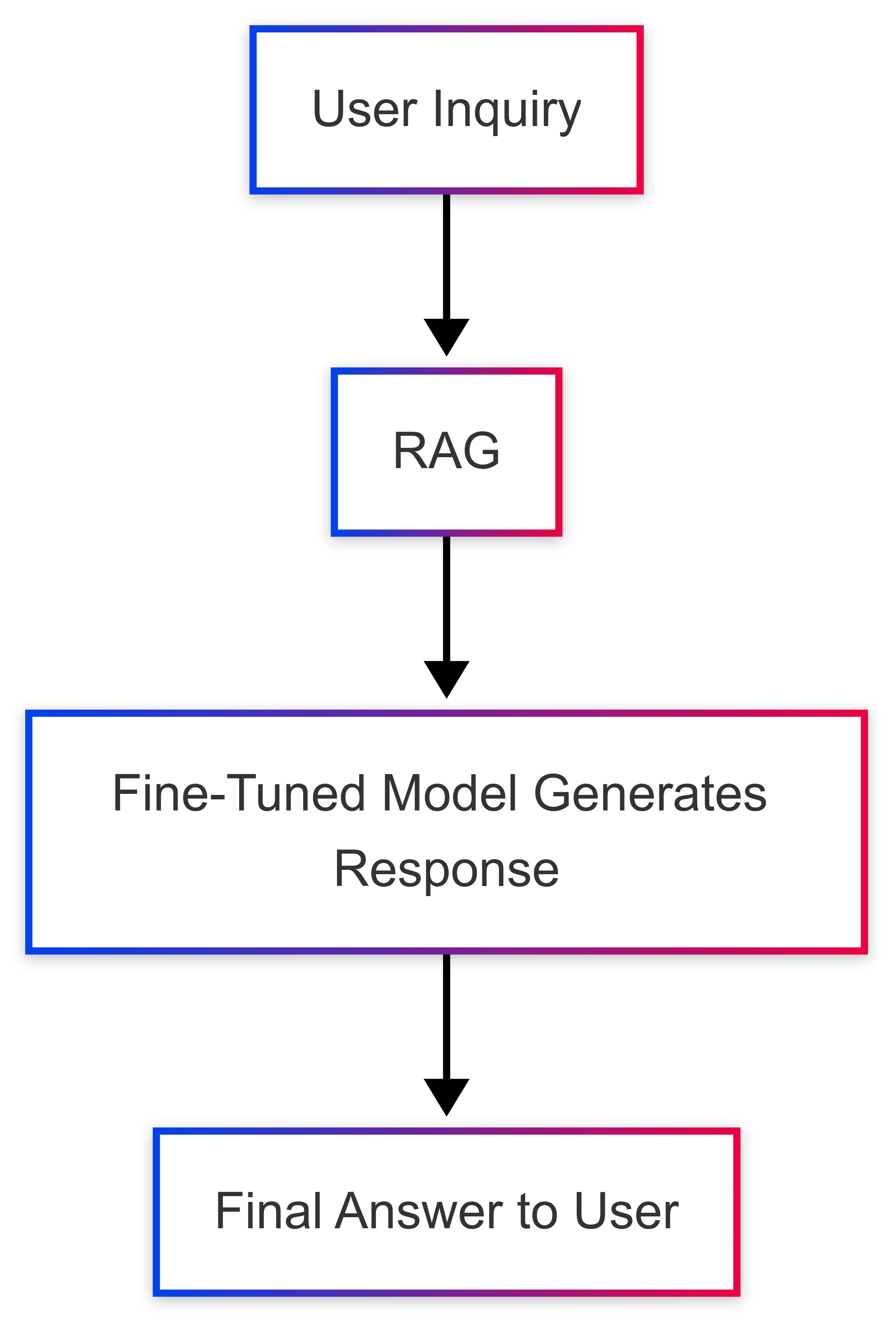} 
    \caption{The flow of the proposed approach}
    \label{fig:enter-label}
\end{figure}

\subsection{Evaluation Metrics}

We use the following metrics to evaluate system performance:

\begin{itemize}
    \item \textbf{Fact Recall (\%)}: Measures the percentage of correct factual statements retrieved and used in the response.
    \item \textbf{Precise Data Recall (\%)}: Focuses on the accuracy of responses that require specific data points such as dates and URLs.
    \item \textbf{User Satisfaction Score}: A subjective rating (1-10) collected from a survey of admissions employees interacting with the system.
\end{itemize}

\subsection{Results and Discussion}

The table below presents a comparison of the four models based on the defined metrics.

\begin{table}[h!] 
    \centering
    \small 
    \caption{Performance Comparison of Response Generation Models}
    \label{tab:performance_comparison}
    \begin{tabular}{|l|c|c|c|} 
        \hline
        \textbf{Model} & \textbf{Fact Recall (\%)} & \textbf{Precise Data Recall (\%)} & \textbf{User Satisfaction (1-10)} \\
        \hline
        Baseline GPT  & 22.3 & 25.6 & 3.2 \\
        RAG Model & 75.1 & 91.4 & 7.5 \\
        Fine-Tuned (No RAG) & 72.7 & 48.3 &  7.9 \\ 
        Fine-Tuned with RAG& 92.7 & 88.3 &  8.9 \\
        \hline
    \end{tabular}
\end{table}

These findings emphasize the importance of combining several approaches: our results show that integrating fine-tuning with RAG significantly improves response accuracy and relevance in university admissions. The fine-tuned RAG model demonstrated strong overall performance, achieving the highest fact recall (92.7\%) and user satisfaction (8.9). While the standard RAG model achieved the highest precise data recall (91.4\%), the fine-tuned RAG model also performed well on this metric (88.3\%) and significantly outperformed the baseline GPT and fine-tuned only models across most metrics. While the fine-tuned only model performed better than the baseline GPT (particularly in fact recall), its precise data recall was significantly lower than the RAG-enhanced approaches, highlighting a key weakness. This highlights the necessity of both retrieval augmentation and domain-specific fine-tuning for optimal results. The superior performance of the combined approach likely stems from RAG's ability to inject precise, up-to-date facts, while fine-tuning equips the model to better understand the context of admissions queries, synthesize retrieved information effectively, and adhere to the institution's specific communication style. The lower precise data recall of the fine-tuned only model suggests that while fine-tuning improves domain understanding, it may struggle to reliably memorize and recall volatile details like specific dates or URLs without a retrieval mechanism. Although highly effective, occasional errors observed in the fine-tuned RAG model involved misinterpreting highly nuanced queries or rare edge cases not well represented in the training data.

\section{Challenges and Future Directions}

\subsection{Current Limitations}

While our approach has shown improvements, some challenges remain. Fine-tuning large models requires significant computational resources, potentially making deployment costly for smaller institutions. Loading a fine-tuned model into memory often necessitates dedicated hardware (e.g., a server with sufficient RAM/GPU memory) for each specific application domain, such as admissions. For example, if the solution is planned to be deployed across different admissions departments such as undergraduate and graduate, two different fine-tuned models might be required as the two different models need to be fine-tuned separately. Beyond cost, limitations include the residual risk of hallucination (though mitigated by RAG), challenges in handling truly ambiguous or novel questions outside the scope of the knowledge base, and the potential for inheriting biases present in the training or source data. Ensuring the system gracefully handles out-of-domain queries remains an area for attention. 

\subsection{Potential Improvements}

Several improvements can be considered to further optimize the model. Firstly, the integration of a dynamic knowledge update mechanism can ensure that the information retrieved is up-to-date. Implementing this would likely require a more complex system, potentially including a user interface for admissions staff to review draft responses and easily update the internal knowledge base used by the RAG component.

\par 
This dynamic updating capability could also mitigate the need for frequent, resource-intensive fine-tuning cycles, especially for rapidly changing information like deadlines or specific policy details. Broad, stable knowledge could be embedded during fine-tuning, while precise, volatile data (like specific dates or URLs) would be best handled by the RAG component. Our results support this, showing RAG's strength in recalling precise data compared to the fine-tuned-only model.

\subsection{Broader Implications}

The successful implementation of fine-tuned RAG models in admissions could inspire broader applications in other domains. Similar techniques could be used in customer service, legal consultations, and healthcare inquiries where accuracy and timely responses are critical. Additionally, improving AI-driven communication tools could reduce the burden on human operators, allowing them to focus on more nuanced cases requiring direct human intervention.

\section{Conclusion}

This paper addressed the challenge faced by university admissions offices in managing high volumes of inquiries while maintaining response quality and timeliness. The core problem lies in providing accurate, context-aware, and personalized information efficiently, which is crucial for prospective student engagement but difficult under heavy workload. We proposed a hybrid solution combining Retrieval-Augmented Generation (RAG) with domain-specific fine-tuning of a large language model. RAG provides access to current, relevant information from institutional knowledge bases, while fine-tuning adapts the model to understand the nuances, terminology, and specific communication style required in the admissions context.

\par 
Our experiments compared this combined approach against baseline models, RAG-only models, and fine-tuned-only models. The results clearly demonstrated the superiority of the integrated fine-tuned RAG system. It achieved the highest overall performance, notably scoring 92.7\% in fact recall and 8.9/10 in user satisfaction, significantly outperforming other configurations. While the RAG-only model excelled in precise data recall (91.4\%), the fine-tuned RAG model was comparable (88.3\%) and offered better contextual understanding and response formulation, as reflected in user satisfaction. The fine-tuned-only model struggled with recalling precise, volatile data (48.3\% recall), highlighting the necessity of RAG for factual grounding. These findings underscore that combining the strengths of RAG (accessing specific facts) and fine-tuning (understanding domain context and style) is crucial for developing effective AI tools for complex, knowledge-intensive tasks like handling admissions inquiries.

\begin{biblio}

\bibitem{cite1}
A. Levin, ``Behavioural responses to customer satisfaction: an empirical study,'' \emph{European Journal of Marketing}, vol. 35, no. 9/10, pp. 1011–1028, 2001. [Online]. Available: \href{https://www.emerald.com/insight/content/doi/10.1108/03090560110388169/full/html}{https://www.emerald.com/insight/content/doi/10.1108/03090560110388169/full/html}.

\bibitem{cite2}
O. Ovadia, M. Brief, M. Mishaeli, and O. Elisha, ``Fine-Tuning or Retrieval? Comparing Knowledge Injection in LLMs,'' in \emph{Proceedings of the 2024 Conference on Empirical Methods in Natural Language Processing (EMNLP 2024)}, 2024, pp. 237–250. DOI: \href{https://doi.org/10.18653/v1/2024.emnlp-main.15}{10.18653/v1/2024.emnlp-main.15}.

\bibitem{cite3}
P. Zhao, H. Zhang, Q. Yu, Z. Wang, Y. Geng, F. Fu, L. Yang, W. Zhang, and B. Cui, ``Retrieval-augmented generation for AI-generated content: A survey,'' \emph{arXiv}, abs/2402.19473, 2024. [Online]. Available: \href{https://doi.org/10.48550/arXiv.2402.19473}{https://doi.org/10.48550/arXiv.2402.19473}.

\bibitem{cite4}
Y. Gao, Y. Xiong, X. Gao, K. Jia, J. Pan, Y. Bi, Y. Dai, J. Sun, and H. Wang, ``Retrieval-augmented generation for large language models: A survey,'' \emph{arXiv}, preprint arXiv:2312.10997, 2023.

\bibitem{cite5}
X. Wang, Z. Wang, X. Gao, F. Zhang, Y. Wu, Z. Xu, T. Shi, Z. Wang, S. Li, Q. Qian, R. Yin, C. Lv, X. Zheng, and X. Huang, ``Searching for Best Practices in Retrieval-Augmented Generation,'' in \emph{Proceedings of the 2024 Conference on Empirical Methods in Natural Language Processing (EMNLP 2024)}, 2024, pp. 17716–17736. DOI: \href{https://doi.org/10.18653/v1/2024.emnlp-main.981}{10.18653/v1/2024.emnlp-main.981}.

\bibitem{cite6}
N. Mecklenburg, Y. Lin, X. Li, D. Holstein, L. Nunes, S. Malvar, B. Silva, R. Chandra, V. Aski, P. K. Reddy Yannam, T. Aktas, and T. Hendry, ``Injecting new knowledge into large language models via supervised fine-tuning,'' \emph{arXiv}, preprint arXiv:2404.00213, 2024. [Online]. Available: \href{https://arxiv.org/pdf/2404.00213}{https://arxiv.org/pdf/2404.00213}.

\bibitem{cite7}
A. Radford, K. Narasimhan, T. Salimans, and I. Sutskever, ``Improving language understanding by generative pre-training,'' OpenAI, 2018. [Online]. Available: \href{https://cdn.openai.com/research-covers/language-unsupervised/language_understanding_paper.pdf}{https://cdn.openai.com/research-covers/language-unsupervised/language\_understanding\_paper.pdf}.

\bibitem{citepre8}
DeepSeek-AI, ``DeepSeek-R1: Incentivizing Reasoning Capability in LLMs via Reinforcement Learning,'' \emph{arXiv}, preprint 	arXiv:2501.12948, 2025. [Online]. Available: \href{https://arxiv.org/abs/2501.12948}{https://arxiv.org/abs/2501.12948}.

\bibitem{cite8}
Gemma Team, ``Gemma 2: Improving open language models at a practical size,'' \emph{arXiv}, preprint arXiv:2408.00118, 2024. [Online]. Available: \href{https://arxiv.org/abs/2408.00118}{https://arxiv.org/abs/2408.00118}.

\end{biblio}

\end{document}